\documentclass[letterpaper,twocolumn]{article}
\usepackage{graphics} % for pdf, bitmapped graphics files
\usepackage{graphicx}
\usepackage{subcaption}
\usepackage{amssymb}  % assumes amsmath package installed
\usepackage{color}
\usepackage{mathtools}
\usepackage{bm}
\usepackage{algorithm}
\usepackage{algpseudocode}
\usepackage{multirow}
\usepackage{caption}
\usepackage{booktabs}

\newcommand{\algName}{P-ALPaCA}
\newcommand{\algNameLong}{Parsimonious ALPaCA}

\newcommand{\citep}[1]{\cite{#1}}
\newcommand{\citet}[1]{\cite{#1}}

\newcommand{\R}{\mathbb{R}}
\newcommand{\xdim}{n_x}

\newcommand{\phidim}{n_\phi}
\newcommand{\nomdim}{n_\text{nom}}
\newcommand{\netdim}{n_\text{net}}

\newcommand{\E}{\mathbb{E}}

\newcommand{\x}{\bm{x}} % state
\newcommand{\ac}{\bm{u}} % action
\newcommand{\feat}{\bm{\phi}} % feature mapping
\newcommand{\weights}{\bm{w}} % network weights
\newcommand{\modelparams}{\bm{k}}
\newcommand{\modelparamsnom}{\bm{k}_{\text{nom}}}
\newcommand{\modelparamsnet}{\bm{k}_{\text{net}}}

\newcommand{\param}{\bm{\theta}} % dynamics latent variable
\newcommand{\paramnom}{\bm{\theta}_{\text{nom}}} 
\newcommand{\paramnet}{\bm{\theta}_{\text{net}}} 
\newcommand{\N}{\mathcal{N}}
\definecolor{darkgreen}{rgb}{0,0.5,0}		%Olivegreen?
\makeatletter
\renewcommand{\paragraph}{%
  \@startsection{paragraph}{4}%
  {\z@}{0.9ex \@plus .5ex \@minus .2ex}{-1em}%
  {\normalfont\normalsize\bfseries}%
}
\makeatother
\usepackage{isairas}
\usepackage{amsmath}

\begin{document}
\title{ADAPTIVE META-LEARNING FOR IDENTIFICATION OF ROVER-TERRAIN DYNAMICS}
\author[1]{Somrita Banerjee}
\author[1]{James Harrison}
\author[2]{P. Michael Furlong}
\author[1]{Marco Pavone}
\affil[1]{Stanford University, 450 Jane Stanford Way, Stanford, CA 94305, USA }
\affil[1]{E-mails: \href{mailto:somrita@stanford.edu}{somrita@stanford.edu},  \href{mailto:jharrison@stanford.edu}{jharrison@stanford.edu},  \href{mailto:pavone@stanford.edu}{pavone@stanford.edu} }
\affil[2]{NASA Ames Research Center, Moffett Field, MS 269-3, CA 94043, USA, E-mail: \href{mailto:padraig.m.furlong@nasa.gov}{padraig.m.furlong@nasa.gov} }

\maketitle
\thispagestyle{empty}

\setlength{\textfloatsep}{3pt plus 1pt minus 1pt}
\setlength{\abovedisplayskip}{4pt} 
\setlength{\abovedisplayshortskip}{2pt}
\setlength{\belowdisplayskip}{4pt} 
\setlength{\belowdisplayshortskip}{2pt}
\addtolength{\belowcaptionskip}{-5mm}
\addtolength{\abovecaptionskip}{-2mm}
%  \setlength\floatsep{1\baselineskip plus 3pt minus 2pt}
%  \setlength\textfloatsep{1\baselineskip plus 3pt minus 2pt}
%  \setlength\intextsep{1\baselineskip plus 3pt minus 2 pt}

% \vspace{5mm}
\section*{Abstract}
% \vspace{5mm}
Rovers require knowledge of terrain to plan trajectories that maximize safety and efficiency. Terrain type classification relies on input from human operators or machine learning-based image classification algorithms. However, high level terrain classification is typically not sufficient to prevent incidents such as rovers becoming unexpectedly stuck in a sand trap; in these situations, online rover-terrain interaction data can be leveraged to accurately predict future dynamics and prevent further damage to the rover. This paper presents a meta-learning-based approach to adapt probabilistic predictions of rover dynamics by augmenting a nominal model affine in parameters with a Bayesian regression algorithm (\algName{}). A regularization scheme is introduced to encourage orthogonality of nominal and learned features, leading to interpretable probabilistic estimates of terrain parameters in varying terrain conditions.

\section{Introduction}
Extraterrestrial rovers must traverse terrain that is often rugged and uncertain in order to achieve mission goals. To ensure a high degree of safety and maximize efficiency, it is crucial for the rover to be equipped with as much knowledge as possible about the terrain it will traverse. In modern rover systems, sources of information include terrain type classification, hazard detection, and excessive slip detection. Terrain type classification relies on input from human operators or machine learning-based image classification algorithms \cite{HigaIwashitaEtAl2019}, \cite{RothrockKennedyEtAl2016}, as well as proposed approaches to classify terrain using vibration measurements \cite{BrooksIagnemma2005}. However, terrain classifications are not sufficient to guarantee safety for path selection. Discrete categories of terrain mask the complexity of vehicle-terrain interactions that are unique to individual vehicles and different environmental conditions. To truly avoid incidents like NASA's Spirit rover becoming stuck in a sand trap \cite{LorenzZimbelman2014}, and Curiosity's wheels being damaged by sharp rocks \cite{ArvidsonIagnemmaEtAl2017}, wheel-terrain dynamics must be considered. Using proprioceptive or visual measurements, unexpected wheel-terrain interaction effects such as excessive slip or sinkage can sometimes be detected as they occur \cite{BiesiadeckiLegerEtAl2007, AngelovaMatthiesEtAl2006, HelmickAngelovaEtAl2009, GonzalezApostolopoulosEtAl2018} but in order to predict and prevent such undesirable wheel-terrain interactions, it is crucial to maintain an adaptive model of the terrain parameters that govern wheel-terrain interaction. Having an estimate of terrain parameters allows the rover to adapt its control and planning strategy to a given terrain \cite{IagnemmaDubowsky2000}, accurately predict the traversability of neighboring terrain regions \cite{Kang2003}, and prevent further damage to the rover.  

\paragraph{Related Work.}
Analytical models governing wheel-terrain interaction were first introduced by Bekker, Wong, and Reece \cite{Bekker1969, Wong2008, WongReece1967} and have since been expanded to incorporate multiple physical effects \cite{DingDengEtAl2015}. However, these models are complex, so efforts to estimate terrain parameters with the limited hardware on-board a rover have motivated approaches using  linear-least squares regression \cite{IagnemmaKangEtAl2004} or a Newton-Raphson method \cite{HutangkabodeeZweiriEtAl2006} on simplified terramechanics models. These approaches require access to sensor information about the rover's forward velocity from an IMU or visual odometry, wheel angular velocity from tachometers, sinkage from camera measurements, and a method for measuring or computing the vertical load and torques on the wheels \cite{IagnemmaKangEtAl2004}. These methods provide good online estimates of terrain parameters on a variety of soils, but cannot model non-linear interactions, and moreover, they do not provide any measure of uncertainty for their estimates.

Since wheel-terrain interaction models have multiple sources of uncertainty, efforts to validate these models against experimental data have relied on fitting stochastic semi-empirical models \cite{BauerLeungEtAl2005}, \cite{Lee2015} which benefit from a measurement of uncertainty for terrain parameter estimates. Such measures of uncertainty can be obtained by modeling the wheel-terrain interactions as a Gaussian Process (GP) \cite{Lee2015} but GPs are limited in their ability to incorporate physical priors, and have high computational cost. Instead, Bayesian neural network models such as ALPaCA \cite{HarrisonSharmaEtAl2018} have been shown to be successful in modeling uncertain dynamics for safety critical systems while remaining computationally efficient \cite{FanNguyenEtAl2019, lew2020safe}. ALPaCA performs Bayesian linear regression on neural network features that are learned via meta-learning (or “learning-to-learn”). By training on a variety of physically plausible simulations, the ALPaCA model is capable of learning expressive features capturing behavior that may be difficult to represent analytically. In this work, we develop a nominal model that relates the dynamics of the rover to the terrain parameters. This model is based on work by Iagnemma et al. \cite{IagnemmaKangEtAl2004} and is linear in terrain parameters. Next, we augment the nominal model with an ALPaCA model \cite{HarrisonSharmaEtAl2018}, to produce a model that yields accurate predictions as well as Bayesian posteriors reflecting prediction uncertainty. A similar approach is employed in \cite{Rogers-MarcovitzGeorgeEtAl2012, Rogers-MarcovitzSeegmillerEtAl2012, SeegmillerRogers-MarcovitzEtAl2013} of relating the pose error to slip estimates, with the differences being that our model estimates terrain parameters instead of slip directly, providing an understanding of the terrain, and our model yields not only predictions but also probabilistic estimates to characterize uncertainty in our predictions.

\paragraph{Problem Formulation.} \label{sec:ProbForm}
We are focused on developing a model to learn, adapt, and predict a rover's dynamics on unknown terrain and provide online estimates of the physical parameters that govern wheel-terrain interactions. The two key parameters of interest are cohesion $c$ and internal friction angle $\phi$ that determine the maximum terrain shear strength \cite{IagnemmaKangEtAl2004} and therefore, a rover's ability to safely traverse the terrain. Since rovers will be encountering novel and unmodeled terrain, it is crucial for safety that our model provide probabilistic predictions and adapt online.

We assume that the sensor suite on-board the rover, including wheel encoders, IMU, and vision system, gives us information about the rover's current position, velocity, slip, and sinkage, though we hope to relax the requirement on knowledge of slip and sinkage in future work.

\paragraph{Contributions.}
The key contributions of this paper are as follows. We develop a framework for wheel-terrain dynamics modeling that is capable of rapid adaptation, by first developing a nominal model that is linear in parameters. In order to capture the non-linear interactions between the rover dynamics and terrain parameters, we augment the linear model with new more complex meta-learned neural network features using a Bayesian regression algorithm (ALPaCA). We present an alternative form of the ALPaCA model, which we name \algName{}, which enables joint inference of the parameters of the ALPaCA model and of the nominal model. We introduce a new regularization scheme to promote orthogonality between the nominal model features and the meta-learned features, providing interpretable terrain parameter estimates. Lastly, we demonstrate our framework on simulations of rocker-bogie dynamics on varying terrains. 

% \paragraph{Organization.}
% In Section \ref{sec:ProbForm}, we formulate our research problem. We present background
% information on meta-learning adaptive control, rover dynamics, and terrain parameter estimation in Section \ref{sec:Background}. In Section \ref{sec:TechApproach} we introduce our approach to estimating terrain parameters and predicting rover dynamics using adaptive meta-learning. Finally, in Section \ref{sec:Experiments}, we demonstrate our framework on simulated rocker-bogie dynamics on varying terrains \footnote{The code for all of our experiments is available at \url{https://github.com/StanfordASL/camelid/tree/cp_prediction}.\sbtodo{rename branch?, make public}} and compare to existing methods of terrain parameter estimation \cite{IagnemmaKangEtAl2004}.

\section{Background}\label{sec:Background}
The dynamics of a rover are different on different terrains. Therefore, predicting rover dynamics is, in its most general form, a system identification problem, where Bayesian regression can be leveraged to produce probabilistic predictions that adapt online. Approaches using Gaussian Process regression have been used for model identification \cite{Lee2015} but are computationally inefficient for large numbers of samples and cannot easily incorporate prior knowledge from the existing models of rover dynamics and wheel-terrain interaction that we discuss in sections \ref{subsec:RoverDyn} and \ref{subsec:WheelTerrInt}. Both these problems are addressed by using a Bayesian meta-learning algorithm, ALPaCA \cite{HarrisonSharmaEtAl2018}, that we discuss in section \ref{subsec:Alpaca} which allows for efficient regression while incorporating prior knowledge.
\subsection{System Identification via Meta-Learned Neural Network}\label{subsec:Alpaca}

% alpaca summary 

Our system identification approach builds upon ALPaCA \cite{HarrisonSharmaEtAl2018}, a Bayesian meta-learning algorithm. This approach separates the online adaptation process---a convex optimization problem that can be solved analytically---from the offline non-convex feature learning process.

The basis of this method is meta-learning, in which it is assumed there exists a distribution over ``tasks''. For example, in system identification, these tasks may correspond to different system dynamics. Meta-learning aims to use data collected from several tasks to improve the efficiency and accuracy of learning in a new task sampled from the same distribution. 

We will write the system dynamics as 
\begin{equation}
    \x_+ = f(\x, \ac; \param) + \epsilon
    \label{eq:basic_dynamics}
\end{equation}
where $\param \sim p(\param)$ corresponds to the task and may be seen as parameters of the environment, and $\epsilon$ is zero-mean Gaussian noise, uncorrelated in time, with covariance $\Sigma_\epsilon$. We write $\x$ and $\ac$ to denote state and action respectively. We assume episodic interaction with the system, where $\param$ is sampled at the beginning of the episode, and held fixed throughout. While this is a somewhat unrealistic assumption in the context of online terramechanics modeling, we will discuss relaxations of this assumption in Section \ref{sec:Conclusions}. Moreover, several works have addressed extensions beyond this assumption that may be applied on top of the methods developed herein \cite{harrison2019continuous, nagabandi2018learning}. We assume $\param$ is not directly observed, and we do not know $f(\cdot,\cdot; \cdot)$. Our goal in the meta-learning problem setting is to learn some approximation of \eqref{eq:basic_dynamics} that is capable of being adapted online. As transitions are observed in an episode, we wish to use experience within an episode to infer $\param$ and improve subsequent predictions.

The approach of ALPaCA is to perform Bayesian linear regression on learned neural network features. The predictive model takes the form
\begin{equation}
    \hat{\x}_+ = K \feat(\x,\ac; \weights) + \epsilon
\end{equation}
where $\feat(\cdot,\cdot; \cdot)$ is a neural network with weights $\weights$, and $K \in \R^{\xdim\times\phidim }$ is a matrix. We write $\hat{\x}_+$ to denote the predicted next state. The matrix $K$ may be seen as the last linear layer of the neural network. Critically, the ALPaCA model maintains a distribution over $K$ reflecting the \textit{epistemic} uncertainty---uncertainty that can be reduced with more data collection. Thus, the ALPaCA model returns a predictive distribution for $\hat{\x}_+$ as opposed to a point prediction. The central idea of ALPaCA models lies in updating the last layer of the network online. This update may be performed by recursive least squares, similar to the Kalman filter. This form of updating is not novel in system identification; indeed, recursive least squares updating on neural network basis functions has been a common approach in nonlinear identification since the 1980s \cite{ljung1983theory}. Where ALPaCA varies from these classical approaches is to train the neural network by backpropagating the training loss through the model update, to ensure the features are broadly useful across multiple tasks and across time. As such, ALPaCA combines ideas from classical adaptive control with ideas from recurrent neural networks. Fundamentally, in the ALPaCA model, the matrix $K$ serves to summarize all information within one episode and account for variations between tasks, whereas the neural network features are learned to be useful across all tasks and are not updated online.

We will now make concrete the above (informal) discussion. We assume access to operational data in a variety of environments, obtained via previous experience or simulation. ALPaCA fixes a matrix normal prior on $K$, which we write 
\begin{equation}
    p(K) = \mathcal{MN}(\bar{K}_0, \Lambda_0^{-1}, \Sigma_\epsilon)
\end{equation}
where $\bar{K}_0$ denotes the prior mean and $\Lambda_0$ is a prior precision (inverse of covariance) matrix. Note that the matrix normal distribution over the matrix $K$ is a normal distribution over the vectorized matrix, with a particular covariance structure. We refer the reader to \cite{HarrisonSharmaEtAl2018} for more details. Let $\mathcal{D}_t = \{(\x_\tau, \ac_\tau, \x_{\tau+1})\}_{\tau=0}^{t-1}$ denote the state/action data observed within an episode at time $t$. Given this data, the posterior over $K$ is 
\begin{equation}
p(K\mid \mathcal{D}_t) = \mathcal{MN}(\bar{K}_t, \Lambda_t^{-1}, \Sigma_\epsilon).    
\end{equation}
where $\bar{K}_t, \Lambda_t^{-1}$ may be computed via recursive least squares through update rules
\begin{align}
    \Lambda_t^{-1} &= \Lambda_{t-1}^{-1} - \frac{1}{1+ \feat_t^T \Lambda_{t-1}^{-1} \feat_t} (\Lambda_{t-1}^{-1} \feat_t) (\Lambda_{t-1}^{-1} \feat_t)^T\\
    Q_t &= \x_t \feat_t^T + Q_{t-1}
\end{align}
where
\begin{equation}
    \bar{K}_t = \Lambda_t^{-1} Q_t
\end{equation}
and where $\feat_t = \feat(\x_{t-1},\ac_{t-1})$. Given this posterior, the posterior predictive distribution may be computed, yielding
\begin{equation}
    \x_{t+1} \sim \N(\bar{K}_t \feat_{t+1}, (\feat_{t+1}^T \Lambda^{-1}_t \feat_{t+1} + 1)\Sigma_\epsilon).
\end{equation}
Given this posterior predictive distribution, we can evaluate the likelihood of the remaining data in the episode (that has not been conditioned upon) to update the neural network weights and the prior.
The prior for the model relies on existing physics models of the rover dynamics and wheel-terrain interactions, which we discuss next.

\begin{figure}[t!]
    \centering
    \includegraphics[width=.9\columnwidth]{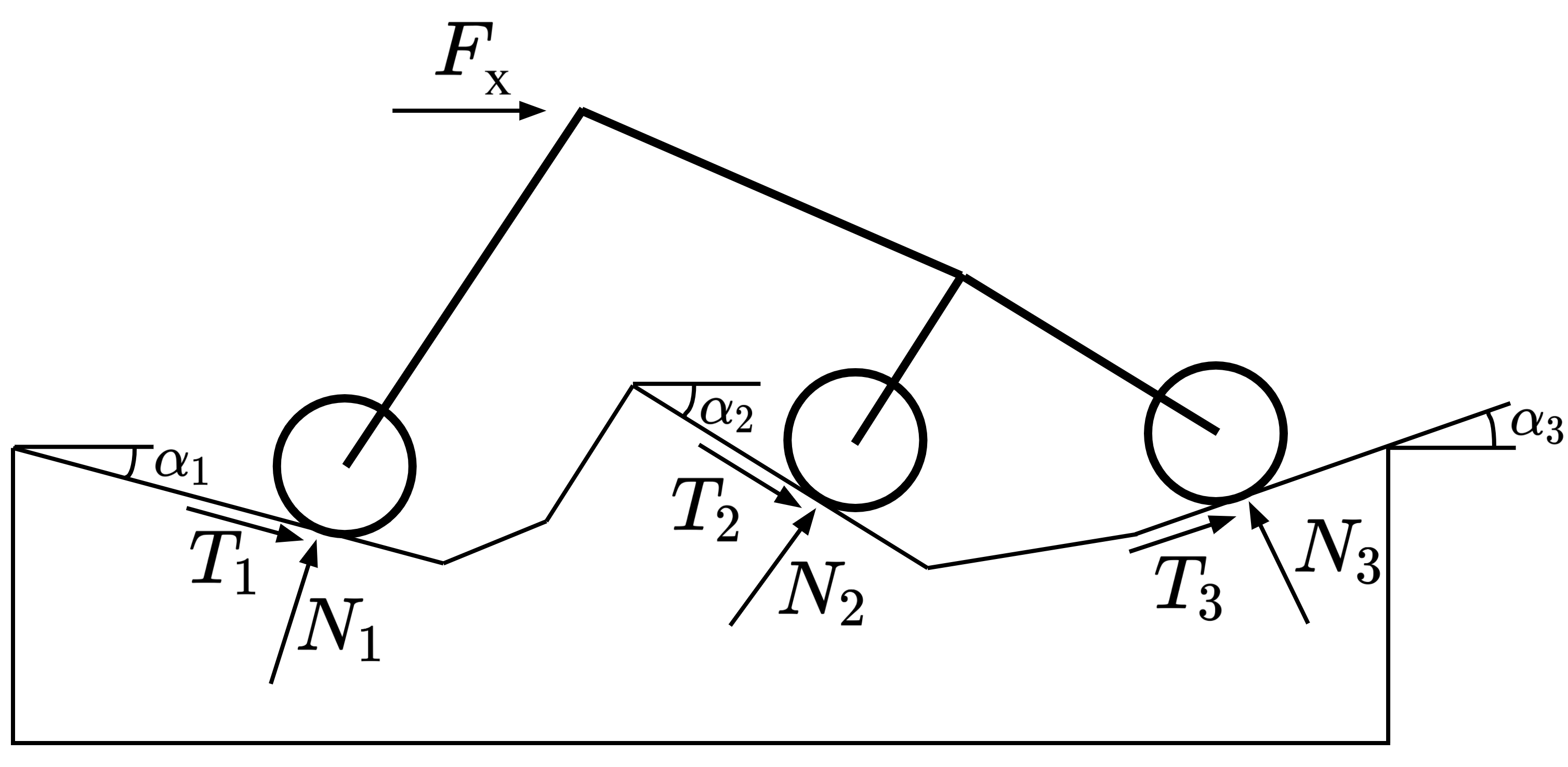}
    \caption{Forces acting on planar rocker-bogie}
    \label{fig:fbd}
\end{figure}

\subsection{Rover Dynamics and Kinematics}\label{subsec:RoverDyn}
 In order to develop a model for rover dynamics on different terrains, we require knowledge of the kinematics of the rover. Though our approach would work on any configuration of the rover, in this work we consider laterally symmetric rocker-bogie motion, i.e. right and left wheels have identical motion and driving torques. This is a justified assumption for low-acceleration motion in a straight-line or very large radius turns.
% and for terrain whose <mean undulation distance> is greater than the width of the rover
With knowledge of the rover kinematics and terrain geometry, the forward net force $F_x$ (see Fig. \ref{fig:fbd}) can be related to the normal $N_i$ and tractive $T_i$ ground reaction forces through 
\begin{equation}
    \begin{aligned}
    F_x &=\bm{S}^T  \begin{bmatrix} N_1& N_2& N_3 & T_1& T_2& T_3 \end{bmatrix}^T \\
    \bm{S} &= \begin{bmatrix} -s_1 & -s_2 & -s_3 & c_1 & c_2 & c_3  \end{bmatrix}^T
    \end{aligned}
    \label{eqn:FxtoNT}
\end{equation}
where $c_i = \cos{\alpha_i}$ and $s_i = \sin{\alpha_i}$ \cite{Hacot1998}.

\subsection{Wheel-Terrain Interactions} \label{subsec:WheelTerrInt}
\begin{figure}[t!]
    \centering
    \includegraphics[width=.68\columnwidth]{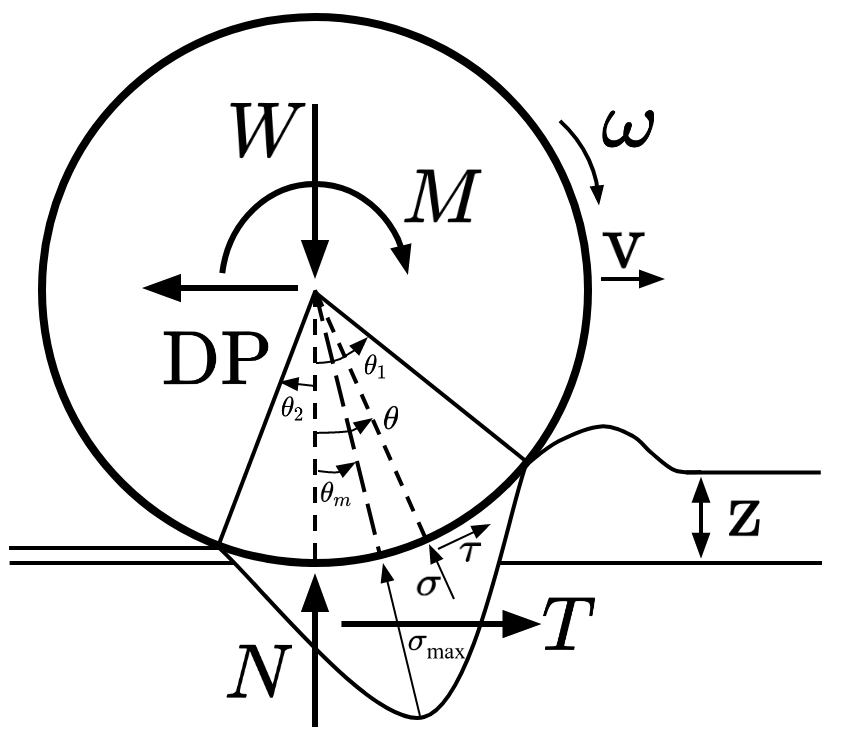}
    \caption{Rigid wheel on deformable terrain.}
    \label{fig:wheel}
\end{figure}

In this section, we describe how the ground reaction forces can be related to the terrain parameters. Fig. \ref{fig:wheel} shows a free body diagram of a rigid wheel
travelling on deformable terrain. For simplicity, the wheel is assumed to be rigid and moving quasi-statically with low acceleration. 
The interaction model for a rigid wheel traversing deformable terrain is obtained by considering the force
and moment balance \cite{IagnemmaKangEtAl2004}
\begin{equation}
\begin{aligned}
    W &= rb\left(\int\limits_{\theta _{1}}^{\theta _{2}} \sigma (\theta)\cos{(\theta)} \, d\theta + \int\limits_{\theta _{1} }^{\theta _{2} } \tau (\theta)\sin(\theta) \, d\theta \right)\\
    {\rm DP} &= rb\left(\int\limits_{\theta _{1}}^{\theta _{2}} \tau (\theta)\cos (\theta) \, d\theta - \int\limits_{\theta _{1}}^{\theta _{2}} \sigma (\theta)\sin (\theta) \, d\theta \right)\\
    M &= r^{2}b\int\limits_{\theta _{1}}^{\theta _{2}} \tau (\theta) \, d\theta
    \label{eqn:WDPM}
\end{aligned}
\end{equation}
where $W$ is vertical load applied to the wheel, ${\rm DP}$ is
drawbar pull applied to the wheel from a suspension
system, $M$ is wheel drive torque produced by an actuator, $\theta$ is the arbitrary angular location of wheel-terrain
contact measured from a vertical axis, $\theta_1$ is the angle
at which the wheel first makes contact with the terrain, $\theta_2$ is the angle at which the wheel loses contact with the terrain, $\sigma$ is radial stress normal to the wheel-terrain contact, $\tau$ is shear stress tangent to the
wheel-terrain contact, $r$ is wheel radius, and $b$ is wheel width. 

In quasi-static equilibrium, the normal force $N$ equals the weight $W$ and the traction force $T$ counteracts the drawbar pull ${\rm DP}$, henceforth $N$ and $T$ will be used in place of $W$ and ${\rm DP}$. By considering linear approximations of the shear and normal stress, Iagnemma et al. \cite{IagnemmaKangEtAl2004} find analytical expressions of these forces
\begin{equation}
\begin{aligned}
    N &= f_0 \left[\sigma _{m} f_1  - \tau _{m} f_2 - c f_3 \right]\\
    T &= f_0 \left[\sigma _{m} f_2  + \tau _{m} f_1 - c f_4 \right]\\
    M &= \frac{r^2 b}{2} \left( \tau_m \theta_1 + c \theta_m \right)
\label{eqn:simpleNTM}
\end{aligned}
\end{equation}
where $\sigma_m$, $\tau_m$ are the maximum radial and shear stresses, $f_{0-4}$ are all functions of $\theta_1$ and $\theta_m$ which is the angle at which maximum stress occurs, which are given by
\begin{equation}
\begin{aligned}
 f_0 &= \frac{rb}{\theta_m(\theta_1-\theta_m)} \\
 f_1 &= {-} \theta_m \cos \theta _{1} + \theta _{1} \cos \theta_m - \theta_1 + \theta_m \\
 f_2 &= \theta_m \sin \theta _{1} - \theta _{1} \sin \theta_m\\
 f_3 &= \theta _{1} \sin \theta_m - \theta_m \sin \theta_m - \theta_m \theta_1 + \theta_m^2 \\
 f_4 &= \theta_{1} \cos \theta_m - \theta_m \cos\theta_m + \theta_m - \theta_1.
\end{aligned}
\end{equation}
The radial and shear stresses themselves are functions of the wheel radius $r$, wheel width $b$, slip $i$, sinkage $z$, the terrain parameters of interest - cohesion $c$ and angle of internal friction $\phi$ - as well as other terrain parameters ($k_c, k_\phi, n, k)$ via 
\begin{equation}
\begin{aligned}
\sigma_m &= \left(\frac{k_{c}}{b} + k_{\phi}\right) (r(\cos \theta_m - \cos \theta _{1}))^{n} \\
\tau_m &= (c + \sigma_m \tan \phi) A\\
A &= 1 - \exp{\left(-\frac{r}{k}[\theta_{1}-\theta_m-(1-i)(\sin\theta_{1}- \sin\theta_m)]\right)}.
\label{eqn:radialandshearstress}
\end{aligned}
\end{equation}
The model exhibits low sensitivity to the other terrain parameters ($k_c, k_\phi, n, k)$ and these are replaced with representative values \cite{WongReece1967}.
The sinkage 
\begin{equation}
\begin{aligned}
z &= r(1 - \cos \theta _{1})
\label{eqn:sinkage}
\end{aligned}
\end{equation}
is determined from the angle of initial contact between the wheel and the terrain $\theta_1$ through wheel geometry. The angle of maximum stress 
\begin{equation}
\begin{aligned}
\theta_m &= (c_1 + i c_2)\theta_1
\label{eqn:th1tothm}
\end{aligned}
\end{equation}
is related to the angle of initial contact $\theta_1$ using terrain parameters $c_1, c_2$. Leveraging these relationships, the forces from Eqs. \ref{eqn:simpleNTM} are expressed as functions of wheel geometry, slip $i$, sinkage $z$, the terrain parameters of interest - cohesion $c$ and angle of internal friction $\phi$ - as well as other terrain parameters. Therefore, with knowledge of the vertical force $N$, torque $M$, sinkage $z$, and slip $i$, the terrain parameters for cohesion $c$ and angle of internal friction $\phi$ are estimated via a least squares regression
\begin{equation}
\begin{aligned}
    \begin{bmatrix} c \\ \tan \phi \end{bmatrix} &= {\bm K}_{2}^{\dagger} {\bm K}_{1}
\end{aligned}
\end{equation}
where ${\bm K}_{1}, {\bm K}_{2}$ are functions of $N, M, i, z$ \cite{IagnemmaKangEtAl2004}.
We build upon this regression model to relate the rover dynamics directly to the terrain parameters for cohesion $c$ and angle of internal friction $\phi$ in the following section.

\section{Technical Approach} \label{sec:TechApproach}
In this section, we build a nominal linear model relating the rover dynamics and terrain parameters. We then incorporate this model in to an ALPaCA-based learning framework. To allow for adaptation of the terrain parameters in the nominal model, we propose an extension to ALPaCA in section \ref{subsec:palpaca} that incorporates models with unknown parameters and performs joint inference on these parameters. In section \ref{subsec:FeatureOrthog}, a regularization scheme is introduced to reduce correlation between nominal model features and meta-learned features, thereby maintaining the interpretability of the terrain parameters.
\subsection{Linear Model Relating Terrain Parameters and Dynamics}\label{subsec:linmodel}
We continue to assume that noisy measurements of slip $i$ and sinkage $z$ are available to the rover, but we relax the requirement in \cite{IagnemmaKangEtAl2004} of measuring the vertical force $N$ and torque $M$ through a force sensor or of computing them on-board. Since force sensors may not be available, and the computation adds overhead, we instead relate the terrain parameter estimates directly to the dynamics of the rover.
Assuming the same linear stress distribution as in Section \ref{subsec:WheelTerrInt}, we can find the ground normal forces $N_i$ and traction forces $T_i$ (for $i = \{1,2,3\}$) as linear in parameters $c$ and $\tan \phi$
\begin{equation}
\begin{aligned}
    N_i &= \bm{Q}_i \begin{bmatrix} c \\ \tan \phi \\ 1 \end{bmatrix} \quad \text{where}\\
    \bm{Q}_i &= \begin{bmatrix} - f_0\cdot \left(f_3 + A f_2\right)& - f_0 \sigma_m A f_2 & f_0 \sigma_m f_1  \end{bmatrix} \\
    T_i &= \bm{R}_i \begin{bmatrix} c \\ \tan \phi \\ 1 \end{bmatrix} \quad \text{where}\\
    \bm{R}_i &= \begin{bmatrix} f_0\cdot \left(-f_4 + A f_1\right)& f_0 \sigma_m A f_1 & f_0 \sigma_m f_2  \end{bmatrix}
\end{aligned}
\label{eq:NTinMatrixForm}
\end{equation}
where $\bm{Q}$ and $\bm{R}$ are 3x3 matrices with non-linear functions of slip $i$ and sinkage $z$. These can then be used to find the forward force using  \eqref{eqn:FxtoNT}, 
\begin{equation}
    F_x = \bm{S}^T \begin{bmatrix} \bm{Q} \\ \bm{R}  \end{bmatrix} \begin{bmatrix} c \\ \tan \phi \\ 1 \end{bmatrix}
\end{equation}
This enables us to use rover dynamics to estimate the terrain parameters or, conversely, to use terrain parameters to predict rover dynamics via
\begin{equation}
\begin{aligned}
\begin{bmatrix}\dot{x} \\ \dot{v} \end{bmatrix} = \begin{bmatrix}v \\ F_x/m \end{bmatrix} 
 = \begin{bmatrix} 0 \quad 0 \quad v \\ \frac{1}{m} \bm{S}^T \begin{bmatrix} \bm{Q} \\ \bm{R} \end{bmatrix} \end{bmatrix}\begin{bmatrix} c \\ \tan \phi \\ 1 \end{bmatrix}
\end{aligned}
\label{eqn:xvdynamics}
\end{equation}
Euler numerical integration is used to yield discrete time dynamics with a chosen time step size of $0.1$ seconds that is sufficiently accurate for the low velocities of the rover. This linear model relating dynamics and terrain parameters is used as a prior for the ALPaCA model described next.

\subsection{\algName{}: Extending ALPaCA with a Parameterized Prior Model}\label{subsec:palpaca}

We now extend the previously proposed ALPaCA model to handle linearly parameterized nominal models. We use the term nominal to refer to a model obtained outside of the meta-learning framework. In this paper, we will consider models derived via simplified dynamics equations, as discussed in the previous subsection. 
Previous work \cite{lew2020safe, harrison2019control} has incorporated prior knowledge into ALPaCA models in the form of a fixed (non-parameterized) nominal model. In particular, previous work has considered system dynamics
\begin{equation}
    \x_+ = f(\x,\ac; \param) + \epsilon = g(\x,\ac) + h(\x,\ac; \param) + \epsilon,
\end{equation}
where $g(\cdot,\cdot)$ is the nominal model, capturing prior knowledge of system dynamics, and $h(\cdot,\cdot; \param)$ is the unknown component of the dynamics that we wish to learn, which varies across episodes/tasks. Previous work has taken $g(\cdot,\cdot)$ as fixed, and parameterized $h(\cdot,\cdot; \cdot)$ with an ALPaCA model
\begin{equation}
h(\x,\ac; K) = K \feat(\x,\ac)
\end{equation}
where $K \in \R^{\xdim \times \phidim}$.

In this work, we investigate integrating nominal models with unknown parameters into the previously developed ALPaCA modeling approach. We investigate models of the form
\begin{equation}
    \x_+ = g(\x,\ac; \paramnom) + h(\x,\ac; \paramnet) + \epsilon,
\end{equation}
where $\paramnom$ denotes parameters of the nominal model and $\paramnet$ denotes (online adapted) parameters of the neural network. This formulation will enable joint inference of unknown parameters from a physical model (e.g. masses) and parameters of the neural network model, which can capture features not appearing in the nominal, simplified physical model. Importantly, joint inference of these parameters is non-trivial, and represents the core machine learning contribution of this paper. 

We will work with a discrete-time nominal model that is affine in the parameters \cite{atkeson1986estimation}, of the form
\begin{equation}
    g(\x,\ac; \modelparamsnom) = \Phi_{\text{nom}}(\x,\ac) \modelparamsnom + \feat_{\text{nom}}(\x,\ac)
\end{equation}
where $\Phi_{\text{nom}}(\x,\ac)$ is a $\xdim \times \nomdim$ matrix of basis functions, $\feat_{\text{nom}}(\x,\ac)$ is a constant (in the parameters) vector of length $\xdim$, and $\modelparamsnom \in \R^{\nomdim}$ is a set of nominal parameters. To fit with this model, we introduce an alternate version of the ALPaCA model of the form
\begin{equation}
    h(\x,\ac; \modelparamsnet) = \Phi_{\text{net}}(\x,\ac) \modelparamsnet
\end{equation}
with $\modelparamsnet \in \R^{\netdim}$. This yields the dynamics model
\begin{equation}
    \x_+ = 
    \begin{bmatrix}
    \Phi_{\text{nom}}(\x,\ac) & \Phi_{\text{net}}(\x,\ac)
    \end{bmatrix}
    \begin{bmatrix}
    \modelparamsnom\\
    \modelparamsnet
    \end{bmatrix} + \feat_{\text{nom}}(\x,\ac) + \epsilon
\end{equation}
for which we write $\modelparams^T = [\modelparamsnom^T, 
    \modelparamsnet^T]^T$ and $\Phi(\x,\ac) = [\Phi_{\text{nom}}(\x,\ac), \Phi_{\text{net}}(\x,\ac)]$. This approach, critically, results in parameters that are shared for each output dimension. Note that $\modelparams \in \R^{\nomdim + \netdim}$; we write $\phidim = \nomdim + \netdim$. In the nominal ALPaCA model, each output dimension is determined by a shared set of basis functions and a unique (for each output dimension) set of parameters. In this model, the parameters are shared for all output dimensions, and each dimension has unique neural network features. Due to the reduced use of adaptive parameters, we refer to this model as \algNameLong{}, or \algName{}.
    
Bayesian inference in this model closely follows from standard results for linear-Gaussian systems (see, e.g., the discussion of Bayesian linear regression in \cite{deisenroth2020mathematics}). We fix a prior $\modelparams_0 \sim \N(\bar{\modelparams}_0, \Lambda_0^{-1})$. Then, the mean and precision (inverse of the covariance) are updated as 
\begin{align}
    \Lambda_t &= \Phi_t^T \Sigma_\epsilon^{-1} \Phi_t + \Lambda_{t-1} \label{eq:precision_update}\\
    \bar{\modelparams}_t &= \Lambda_t^{-1}(\Lambda_{t-1} \bar{\modelparams}_{t-1} + \Phi_t^T \Sigma_\epsilon^{-1} (\x_{t} - \feat_{t}))
\end{align}
where $\Phi_t = \Phi(\x_{t-1},\ac_{t-1})$ and $\feat_{t} = \feat_{\text{nom}}(\x_{t-1},\ac_{t-1})$. In the standard ALPaCA formulation the Woodbury identity is applied to yield an efficient update rule. Applying the same approach here does not yield computational efficiency improvements, as we do not avoid a matrix inverse in the update. Thus, we maintain the natural parameters of the multivariate Gaussian during updating via
\begin{equation}
    \bm{q}_t = \bm{q}_{t-1} + \Phi_t^T \Sigma_\epsilon^{-1} (\x_{t} - \feat_{t})
\end{equation}
and the precision update \eqref{eq:precision_update}. To perform prediction, we may compute $\bar{\modelparams}_t = \Lambda_t^{-1} \bm{q}_t$. Then, the posterior predictive distribution is 
\begin{equation}
    \x_{t+1} = \N(\Phi_{t+1} \bar{\modelparams}_t + \feat_{t+1},\Phi_{t+1} \Lambda^{-1}_t \Phi_{t+1}^T + \Sigma_\epsilon )
\end{equation}
which may be derived based on the standard equations for mean and variance of a random variable under linear transformations. Given this posterior predictive, the likelihood may be computed as in the standard ALPaCA meta-learning algorithm, and used to train the neural network features as well as the prior over $\modelparams$.
Due to space constraints we do not outline the full training algorithm in this paper; we refer to \cite{HarrisonSharmaEtAl2018} for more details.

\begin{figure*}[t!]
\centering
\includegraphics[width=0.95\textwidth]{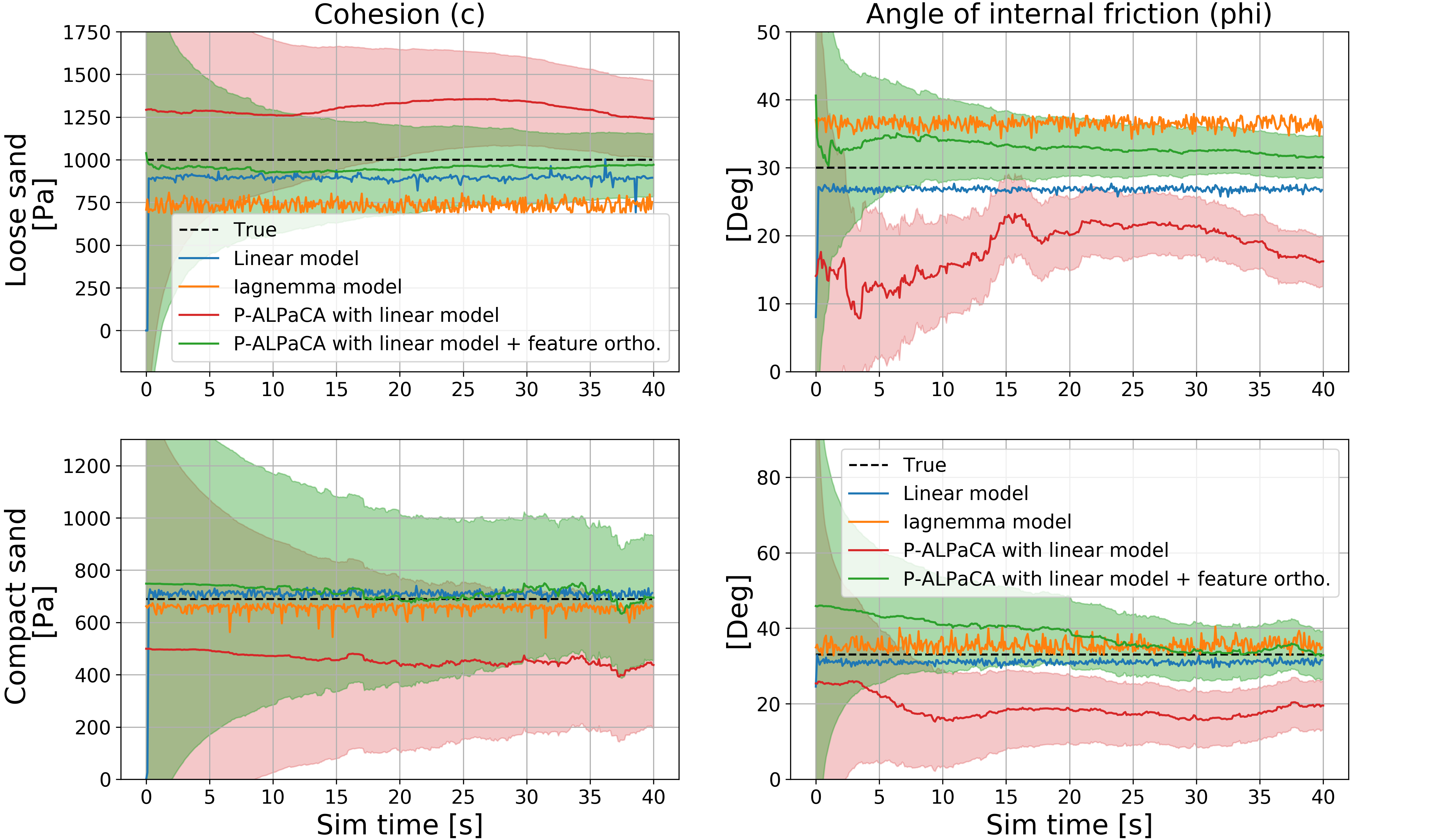}
\caption{Estimate of terrain parameters on loose and compact sand}
\label{fig:est_param}
\end{figure*}

\subsection{Interpretability via Feature Orthogonality}\label{subsec:FeatureOrthog}

We have so far presented an approach to combine a ``grey-box'' nominal system model, derived via simplified physics of the rover system, with an entirely black-box neural network model without physical interpretation. If our only objective is accurate state prediction, then our approach so far is sufficient. More likely, however, the physical parameters of the grey-box model provide information that is of use to system operators, scientists, or elsewhere in the planning and control architecture. For example, inferred terramechanics properties may be critical to identify regions as safe or unsafe for traversal. Thus, it is necessary that our added black-box model not result in poor estimates of the physical parameters in the grey-box model. Such an outcome is not guaranteed; collinearity between features in linear regression results in weights that are highly unstable under minor perturbations \cite{greene2003econometric}. This collinearity may result in the parameters of the neural network model being highly correlated with the grey-box model, resulting in misleading estimates of the physical parameters. To reduce this collinearity, we add a regularization term to the loss function to promote feature orthogonality. 

Let $\feat^T_{i,t}$ denote the $i$'th row of $\Phi_t$\footnote{Note the notation conflict with the previous subsection; we will ignore the constant (with respect to parameters) term in the nominal model in this section for ease of presentation. }. Our orthogonality regularization takes the form 
\begin{equation}
     \sum_{\text{tasks}} \sum_{i=1}^{\xdim} \left\|I - \frac{1}{T} \sum_{t=0}^T \feat_{i,t} \feat^T_{i,t} \right\|^2_F
\end{equation}
where $\| \cdot \|_F$ denotes the Frobenius norm and $T$ is the length of the episode. There are several things to note about the above. The norm of the innermost sum over time is a Monte Carlo approximation of the constraint
\begin{equation}
    \E_{\x\sim\mu, \ac \sim\pi}\left[\feat_i(\x,\ac) \feat_i(\x,\ac)^T \right] = I
\end{equation}
where $\pi$ is the controller used for data collection and $\mu$ is the occupancy measure (i.e. the probability distribution over states) induced by the system dynamics and the controller. Thus, the addition of this regularizer may be interpreted as a Lagrangian relaxation of the neural network training loss with a constraint requiring the basis functions to be orthogonal, as 
\begin{equation}
    \E_{\x\sim\mu, \ac \sim\pi}\left[\feat_{ij}(\x,\ac) \feat_{ik}(\x,\ac)^T \right] = 0 \,\,\, \text{for} \,\,\, j \neq k
\end{equation}
where $\feat_{ij}$ denotes the $j$'th entry of $\feat_{i}$. As orthogonality of features implies no multicollinearity, this regularization term will minimize the effect and thus improve interpretability of the model. 

\section{Experiments} \label{sec:Experiments}

Simulations were conducted of a rover in rocker-bogie configuration traveling over two types of terrain - compact sand and loose sand - using equations for the wheel-terrain interaction and terrain parameters from Wong et al. \cite{WongReece1967}. For a given rover body position, the rover dynamics were simulated by computing the ground reaction forces for each wheel from applied torques and matching against the integrals from Wong's model which were solved using Simpson's rule. The ground reaction forces were then used to calculate the slip and sinkage for each wheel, which, along with rover position and velocity, were the inputs to the models. 
Rover kinematics was used to compute the net forward force and resultant change in rover velocity. The purpose of the simulations was to compare the performance of the nominal linear model against the model augmented with \algName{} features and analyze the effect of enforcing feature orthogonality. 

The simulation\footnote{The code for all of our experiments is available at \url{https://github.com/StanfordASL/rover-meta-learning}.} was written in Python and the machine learning models were implemented using the PyTorch library \cite{PaszkeGrossEtAl2017}. Each of the models consisted of two hidden layers of size 128 and were trained for 1000 iterations in batches of 20 sampled dynamics transitions each to minimize the negative log likelihood of the posterior predictive distributions. The first model was \algName{} without any added features, the second model was \algName{} with two added nominal features corresponding to the linear model features for velocity from Eq. \ref{eqn:xvdynamics}, and the third model was \algName{} with two added nominal features as well as an additional regularization term to promote orthogonality between features as discussed in Section \ref{subsec:FeatureOrthog}. The \algName{} models with nominal features had an additional loss term to encourage the parameters associated with cohesion and internal friction angles to be positive. 

The state consisted of velocities between 0 to 1 m/s and the control actions were applied wheel torques between 0 and 5 Nm, wheel slips between 0 and 1, as well as sinkage between 0 and 0.1 m. The rover mass was 10 kg, the wheel radius was 0.4 m, and wheel width was 0.1 m. During training, the parameter for cohesion was randomly selected for each batch between 0.5 and 5 kPa, except the band between 0.7 and 1.3 kPa which are the typical values of cohesion for loose and compact sand \cite{WongReece1967} and were reserved for testing. The internal friction angle was randomly selected from the range 0 to 60 degrees, excluding the range from 20 to 45 degrees for testing. The other terrain parameters were held fixed at the values for loose sand ($n = 1.1$, $k_c = 0.9$ kPa, $k_\phi = 1523.4$ kPa, $k = 0.025$, $c_1 = 0.18$, $c_2 = 0.32$) and compact sand ($n = 0.47$, $k_c = 0.9$ kPa, $k_\phi = 1523.4$ kPa, $k = 0.038$, $c_1 = 0.43$, $c_2 = 0.32$) \cite{WongReece1967} during training and perturbed by 5\% during testing to ensure robustness.

\begin{figure}[t!]
\centering
\includegraphics[width=0.47\textwidth]{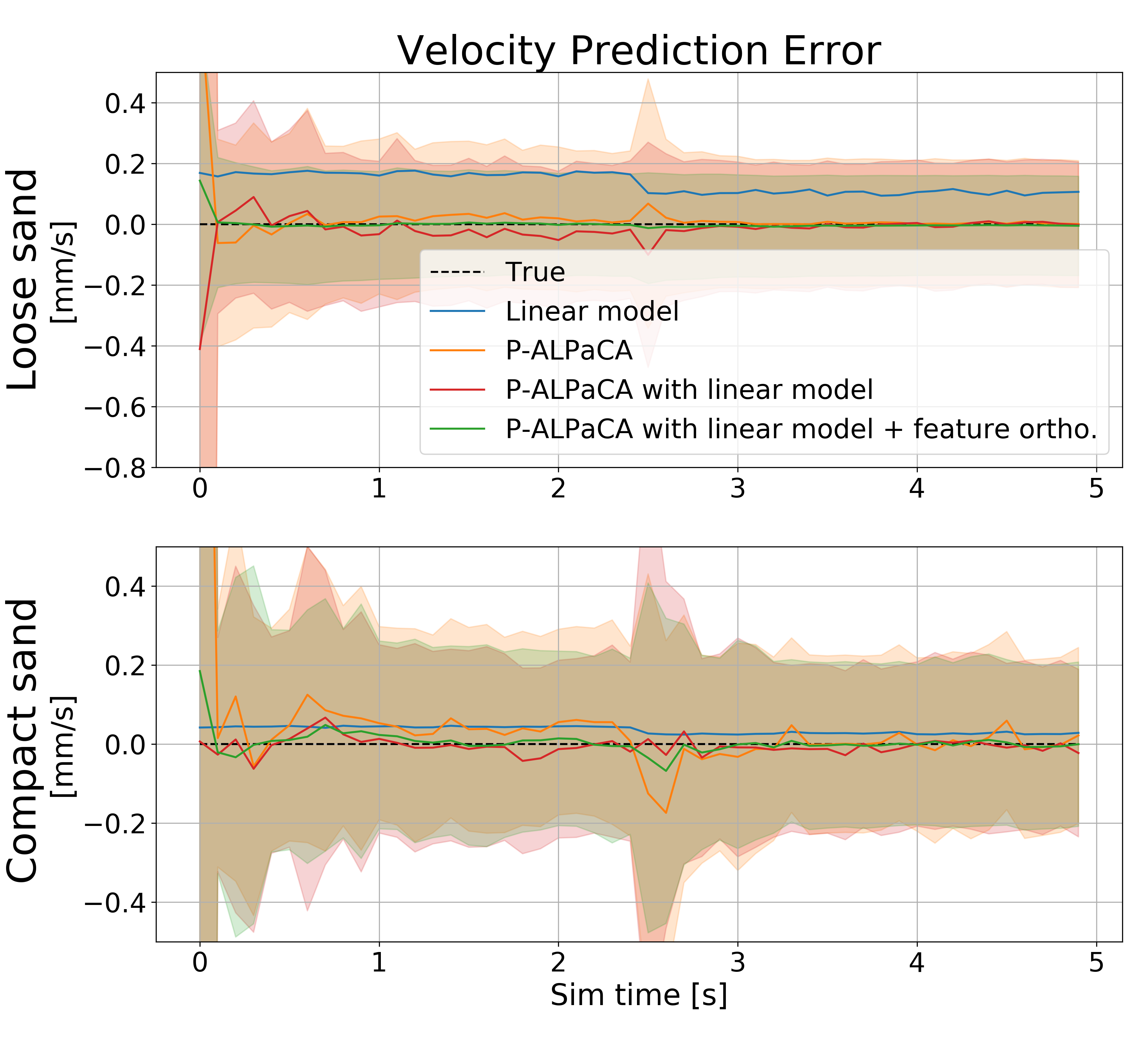}
\caption{Prediction error for velocity}
\label{fig:pred_dyn}
\end{figure}

First, we compare the ability of the models to predict the dynamics of the rover given an estimate of the terrain parameters. As shown in Fig. \ref{fig:pred_dyn}, the linear model as well as all the \algName{} models have low prediction error for velocity of the rover, with the largest being the error for the linear model on loose sand, which is less than 5\%.

Second, we compare the terrain parameter estimates provided by the models using the dynamics of the rover, as well as the estimates provided from Iagnemma et al.'s model \cite{IagnemmaKangEtAl2004}. The ground truth of vertical load and wheel torques are passed from the simulator to Iagnemma's model, while the other models utilize the position and velocity of the rover directly to make predictions. As seen in Fig. \ref{fig:est_param}, both the linear model and Iagnemma's model provide good estimates for cohesion and angle of internal friction. The higher error for Iagnemma's model on loose sand is due to their assumption that the angle of maximum shear stress is located halfway from the angle of initial contact, which becomes a looser assumption for higher cohesion terrains. While both \algName{} models augmented with the linear model provide uncertainty estimates, the terrain parameter estimate shows better convergence when orthogonality of the features is enforced explicitly through the regularization term discussed in Section \ref{subsec:FeatureOrthog}.

\begin{figure}[t!]
\centering
\includegraphics[width=0.96\columnwidth]{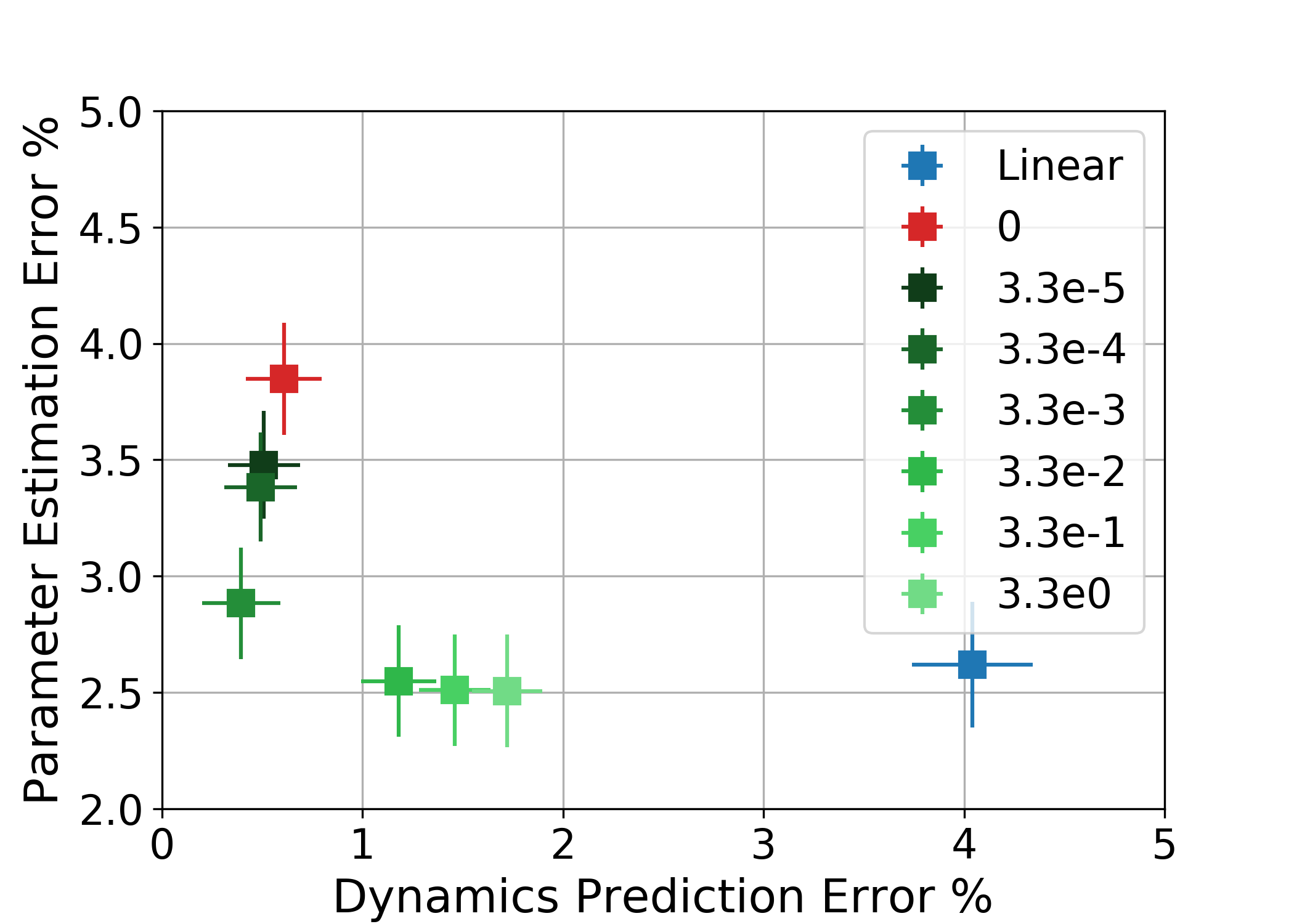}
\caption{Effect of regularization strength}
\label{fig:orthog}
\end{figure}

The addition of the regularization scheme results in interpretability of the meta-learned features. \algName{} with orthogonal regularization not only provides dynamics predictions but also an estimate of the terrain parameters used in the model, aiding in interpretability of the model. 
Models with orthogonal regularization produce lower parameter estimation error than models without the regularization and lower dynamics prediction error than the nominal linear model. Increasing the orthogonal regularization strength, i.e. increasing the weight on the orthogonal loss, results in a trade-off between lower parameter estimation error and higher dynamics prediction error, as seen in Fig. \ref{fig:orthog}.

\section{Discussion and Conclusions}\label{sec:Conclusions}

% contributions 
Our approach of augmenting a nominal model affine in parameters with meta-learned neural network features from ALPaCA leverages Bayesian regression to adapt to rover dynamics. We maintain interpretability of the predictions by encouraging orthogonality of the nominal model features and the meta-learned features, resulting in good terrain parameter estimates. Our experimental results demonstrate that the combination of the grey-box nominal model with a properly regularized black-box neural network model results in achieving the best of predictive accuracy and accuracy in parameter inference. Thus, our proposed approach is a highly flexible and performant system identification framework that is capable of application to many systems. 

Developing terrain parameter estimation models that quickly adapt to novel terrains is a step towards greater autonomous operations of rovers, while increasing safety and efficiency. Directions for future work include incorporating more sources of information such as vision systems, detecting both sharp and gradual changes in terrain types, testing on physical prototypes, incorporating bulldozing, high slope, and multipass effects, and developing combined parameter estimation and adaptive slip prediction models.

\subsection*{Acknowledgement}
Somrita Banerjee and James Harrison were supported by the Stanford Graduate Fellowship. The authors were partially supported by an Early Stage Innovations grant from NASA's Space Technology Research Grants Program. The authors wish to thank Abhishek Cauligi, Thomas Lew, and Apoorva Sharma for their helpful feedback.

% Put in current ASL bib and main.bib

\renewcommand{\baselinestretch}{.95}

\nocite{article_example,inproceedings_example,misc_example,phdthesis_example,book_example}
\bibliographystyle{isairas}
{\small 
    \bibliography{main, ASL_papers, cites}
}

\end{document}